\documentclass[11pt]{article} 

\usepackage[utf8]{inputenc} 

\usepackage{geometry} 
\usepackage{graphicx} 

\usepackage{booktabs} 
\usepackage{array} 
\usepackage{paralist} 
\usepackage{verbatim} 
\usepackage{caption}
\usepackage{subcaption} 
\usepackage{float}
\usepackage{wrapfig,lipsum,booktabs}

\usepackage{fancyhdr} 
\usepackage{wrapfig}

\usepackage{sectsty}
\allsectionsfont{\sffamily\mdseries\upshape} 

\usepackage[
backend=bibtex,
maxalphanames=4,
style=numeric,
maxbibnames=99,
sorting=none,
]{biblatex}

\usepackage[nottoc,notlof,notlot]{tocbibind} 
\usepackage[titles,subfigure]{tocloft} 

\graphicspath{ {Figures/} }

\title{A Novel Review of Stability Techniques for Improved Privacy-Preserving Machine Learning} 
\author{Coleman DuPlessie$^*$, Aidan Gao$^*$} 
\footnotetext{Equal contribution}
\date{} 

\begin{document}
\maketitle

\begin{abstract}

Machine learning models have recently enjoyed a significant increase in size and popularity. However, this growth has created concerns about dataset privacy. To counteract data leakage, various privacy frameworks guarantee that the output of machine learning models does not compromise their training data. However, this privatization comes at a cost by adding random noise to the training process, which reduces model performance. By making models more resistant to small changes in input and thus more stable, the necessary amount of noise can be decreased while still protecting privacy. This paper investigates various techniques to enhance stability, thereby minimizing the negative effects of privatization in machine learning.
\end{abstract}




\section{Introduction}

Data, especially private data, has become increasingly valuable in the modern era. From hospital records to personal search histories, the increased collection and use of private data means that data analysis conducted on these data sets must protect sensitive information about individuals. Without this protection, a leak of sensitive information could easily have lasting consequences for an individual, even from seemingly innocuous data like a photo. The rise of machine learning has exacerbated these concerns even further due to its need for specific and abundant data to produce accurate predictions. This large amount of required data and machine learning models' tendency to memorize specific yet unnecessary information, such as specific IP addresses during text responses, makes private machine learning especially important\cite{TheSecretSharer}.

 Releasing models trained on private data can allow observers to extract private and personal information, compromising the algorithm's utility. One area where this is especially important is Large Language Models (LLMs), which can often be tricked into repeating portions of their training data verbatim\cite{carlini2021extracting}. However, since the process of training a machine learning model is simply a (complex) algorithm that takes in training data and outputs a trained model, it is theoretically possible to privatize the machine learning process. Different mathematical frameworks underlie this privatization process, but implementing these frameworks always involves adding random noise to some steps of the training process. The traditional method, known as differential privacy (DP), adds noise to the model's gradient during training, using the sensitivity of the training process to calculate how much noise is necessary in the theoretical, adversarial worst case. We, however, base our work on the assumption that noise will be added to parameters of the trained model, either at the end of training or at regular intervals throughout, and use a different mathematical framework, Probably Approximately Correct (PAC) privacy, that, although weaker than differential privacy in the worst case, is more flexible and still protects privacy in the majority of cases\cite{PACPrivacy}.
 \pagebreak 
 
These frameworks, differential privacy and PAC privacy, prevent inference about characteristics of any specific data while still allowing general inferences over the whole dataset to be made\cite{Dwork_Roth_2014}. By ensuring that a model follows the properties of either of these frameworks, we guarantee that including someone's personal data in the training dataset will not result in harm, regardless of what happens to the trained model. However, the random noise necessary to ensure privacy often produces a severe loss in accuracy as a tradeoff\cite{OurData}. Because the PAC framework bases the magnitude of the required noise on stability, we can ensure that privatization is minimally harmful by making the training process maximally stable. A stable model is one for which adding or removing a small amount of training data will only result in a slight change in the final model and, therefore, allows us to lower the magnitude of random noise needed to protect against edge cases. By ensuring that a model naturally avoids hinging on relatively few pieces of training data, the noise required to privatize it consequently lowers, allowing for a more robust model while maintaining the privacy standard. This paper investigates and compares several techniques to improve model stability and noise generation. The most potent techniques have been data transformations, whole-batch gradient clipping, regularization, and group-sample gradient clipping. In our results, we develop significant improvements in stability in tests with both Resnet20s and simple linear models.
 
\section{Background}

\subsection{Privacy Mechanisms}
Practically, private systems are created by adding random noise at some stage. Because post-processing of privatized information (such as inference run on a model trained with DP or PAC privacy) will always preserve the original privacy guarantee, noise can be added at any point in a machine learning model's training cycle from the training dataset to the finished model. The current state-of-the-art uses differential privacy to add noise to the gradient during stochastic gradient descent. Since we are only adding a finite amount of noise, however, to have a true guarantee of privacy for any individual piece of data, the state-of-the-art must guarantee that any given sample, no matter how extreme or unrealistic, can only change the overall gradient by a limited amount. This is done by per-sample gradient clipping, in which any samples whose gradients exceed a particular ceiling have their gradients reduced in magnitude before all gradients in a batch are averaged together. By doing this, we can guarantee that changing one sample will cause a change in the overall gradient of at most twice the clipping value (in the case where the before-and-after gradient of the sample is both strong and diametrically opposed). This can, of course, be a problem since it disrupts model behavior and prevents it from emphasizing small samples, even when those samples should be weighted more heavily (e.g., if a model is performing disproportionately poorly on one subset of input data, it ought to `pay more attention' to that data).

The main problem of private machine learning is that it comes with large performance costs. Both adding random noise to the gradient and artificially clipping individual samples' gradients interfere with gradient descent, leading to less accurate models that achieve lower top accuracy. Additionally, the fact that differential privacy deals explicitly with the gradients of individual samples means that many layers that take other samples in the same batch into account, such as the widely used BatchNorm layers, must be either removed or replaced by inferior substitutes since they allow individual samples to influence not just their own gradients, but also potentially the gradients of all other samples in their batch.

\subsection{Stability}
Stability measures a model's resistance to small changes in input\cite{stability}. While privacy systems like DP and PAC privacy describe the criteria necessary to make a model private, stability is a measurement of machine learning models' reactivity to small changes in input. Models with higher stability need less noise to be made private for any given degree of privacy, making stability a key factor to be optimized. With past research having found that stochastic gradient descent is generalizable and stable, current work will focus on model specifics and structure and modifications to the basic pattern of SGD, aiming to increase stability while not sacrificing accuracy\cite{SGDstability}.

Traditional differential privacy requires that we guarantee that a model is resistant to arbitrary, adversarial changes in input. It requires a low sensitivity instead of a high stability. However, PAC privacy enables us to determine the proper amount of noise necessary to privatize a model with a given level of stability\cite{PACPrivacy}. Since our experiments focused on optimizing stability rather than sensitivity, we use PAC privacy as our privacy framework throughout this paper. As a benchmark, we should expect the divergence between two models trained on nearly-identical datasets that differ in only one datapoint to, in the best case, scale with $\frac{1}{n}$, where $n$ is the size of the dataset.

\section{Methods}
\subsection{Constants}
\subsubsection{Models}

Since our experimentation is entirely conducted on the CIFAR-10 dataset, we primarily use various convolutional neural networks (CNNs). While an ECNN and VGG 19 structure were briefly used along with smaller custom CNNs, the Resnet20 architecture was decided upon as the main model for most experiments due to its relatively small size and high accuracy on the CIFAR-10 dataset, at a top accuracy of 90.4 percent\cite{Resnet}. 

In some experiments, we instead use a single-layer linear network, chosen because of its convex nature and small size, allowing for complete stabilization in the optimal case. The smaller size also allows for a defined impact from noise and reproducible trials due to its highly deterministic nature. We have also experimented with simple, non-ML linear regressions calculated using least-squares as a benchmark.

\subsubsection{Dataset}

We conduct all experiments on the CIFAR-10 dataset, which consists of colored 32x32 images classified into 10 classes. The dataset contains 50,000 RGB images for training and 10,000 images for testing\cite{CIFAR}.

\begin{wrapfigure}{r}{8.1cm}
\begin{center}
\includegraphics[width=8cm]{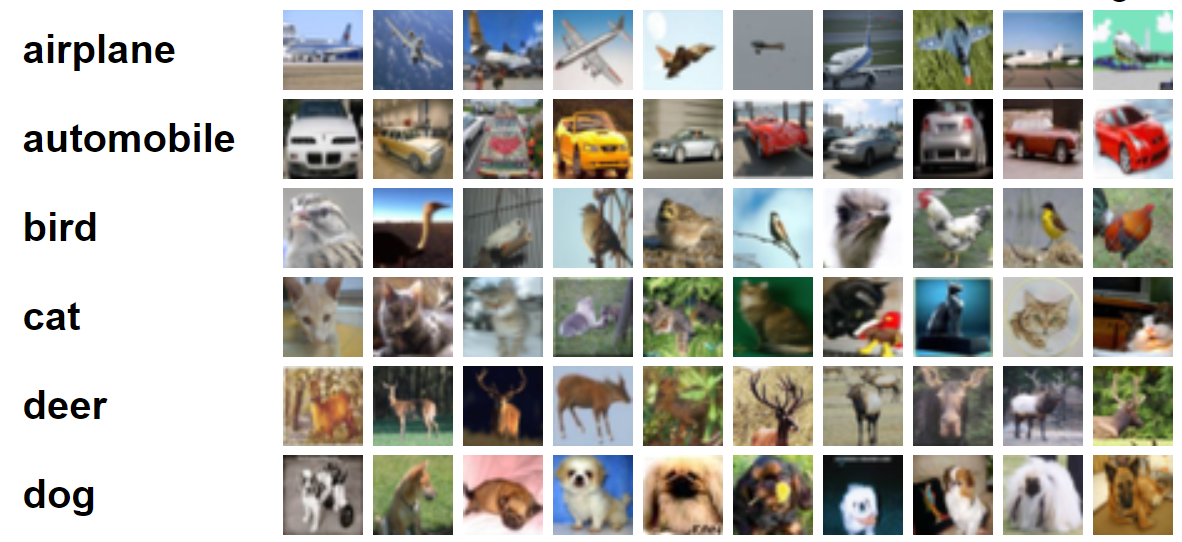}
\caption{Some samples from the CIFAR-10 dataset}
\end{center}
\end{wrapfigure}

Standard normalization values were applied to the dataset, and a random crop with a padding value of 4 and a final size of 32x32, along with a random horizontal flip, was used as data augmentation. These prevented model overfitting and further tested stability by creating more distinctions between random subsets and model trials.

In addition to these transforms, a random color jitter and random augment were used in some experiments to prevent overfitting. 

\subsubsection{Stability Calculations}

To calculate model stability, we train several copies of the same model (all initialized with the same random parameters) on different random subsets of CIFAR-10. Once we have several different trained models, we take the average of their parameters. We then subtract the average parameters from each individual model and calculate the l2 norm of the remainder, averaging all l2 norms from each model as the final value. This measurement is coined the `deviation l2 norm,' or simply `deviation.' We also calculate the average l2 norm of the individual models, which allows us to express the deviation as a fraction of the average model size. We refer to this second quantity as `percent l2 norm deviation' or `percent deviation.' In other experiments, we instead stack the parameters of the $n$ models into a $n\cdot m$ matrix, where $m$ is the number of parameters, then calculate the eigenvalue decomposition of the product between it and its transpose\cite{eigendecomp}. The resulting square root sum of the eigenvalues can then be taken, allowing another metric to calculate the stability for a set of models. Functionally, the eigenvalues produced should be similar to the deviation but provide the benefit of optimal noise generation along the diagonal matrix with the eigenvalues, as noise can be added proportionally to the vector subspaces with the largest variance.

\subsection{Baselines}

In all experiments, we trained several copies of the same model, each initialized with the same weights to eliminate randomness. While pre-trained models are used as starting points for accuracy in a few experiments, most experiments had the model initialized with random noise. If a pretrained model is used, it is trained on 5,000 random points from CIFAR-10, representing a hypothetical public dataset that can be collected without privacy concerns. Given that all information potentially revealed from the model will already be available, there is no need for the pre-trained model to be private, which creates a good baseline accuracy for a model trained on private information (in this case, the rest of CIFAR-10) to build off of. 

In most experiments, we use full-batch gradient descent, which utilizes the entire subset of training data as the batch size of the SGD optimizer. Since setting the batch size to the dataset size makes the optimizer theoretically deterministic, under ideal conditions, two models trained using the same conditions would be identical (although there is a slight difference due to some low-level functions and approximations made by PyTorch being nondeterministic, this difference is insignificant). By using full batch gradient descent, an almost deterministic process, we can confidently measure small changes in outcome, such as those caused by removing or adding 10 data points to the training set, that would normally be obscured by the inherent randomness of stochastic gradient descent.  This allows us to establish the baseline divergence caused by the training process and gives us a benchmark to measure against when utilizing random subsets of data.

\subsection{Resnet20}
Four experiments to improve stability were conducted on the Resnet20, including layer freezing, pruning, gradient clipping, and developing a tree net.

The Resnet20 model was first trained on a random subsample of 5,000 images from CIFAR-10, following the pretrained start to conduct layer freezing. After pretraining on 5,000 samples, only the last linear layer was unfrozen, while all convolutional layers were kept frozen. The model was then trained on another 25,000 random samples over 100 epochs to observe changes in accuracy and stability. 

Built-in PyTorch functions were used to conduct pruning and gradient clipping. Models were pruned based on their l1 norm, maintaining the more extreme parameters while reducing the overall number of parameters. Additionally, gradients were clipped based on the l2 norm. These two values were fine-tuned to find optimal values and measure changes in accuracy and divergence from 100 training epochs using a pretrained base described previously.
 
To counteract the increased divergence of larger models due to variation from the larger number of parameters, several smaller models were utilized to maintain accuracy while improving stability. A binary tree structure was used to combine the smaller models, hence the name of a tree net. By classifying the classes within the CIFAR-10 dataset into several binary classification tasks, it is possible to train several lower-scale binary classification models on different subsets and then use a binary decision tree where the output of the first model dictates which models on the second layer are used, and so forth. The binary tree used was built with Resnet8 blocks, which suffer an accuracy decrease of around 1-2 percent for each task but use only a quarter of the parameters of a Resnet20.

To test the varying effectiveness of this structure, two different architectures were tested against the Resnet20 used for previous testing: a two-layer tree net and a four-layer tree net. The two-layer tree net uses a Resnet8 block as a binary classification between vehicles and animals, then two separate Resnet20s as classifiers for the animals and vehicles. The four-layer tree net utilizes a full binary net structure, using Resnet 8s as classifiers following the classification schema in the picture below:

\begin{figure}[H]
\begin{center}
\includegraphics[width=15cm]{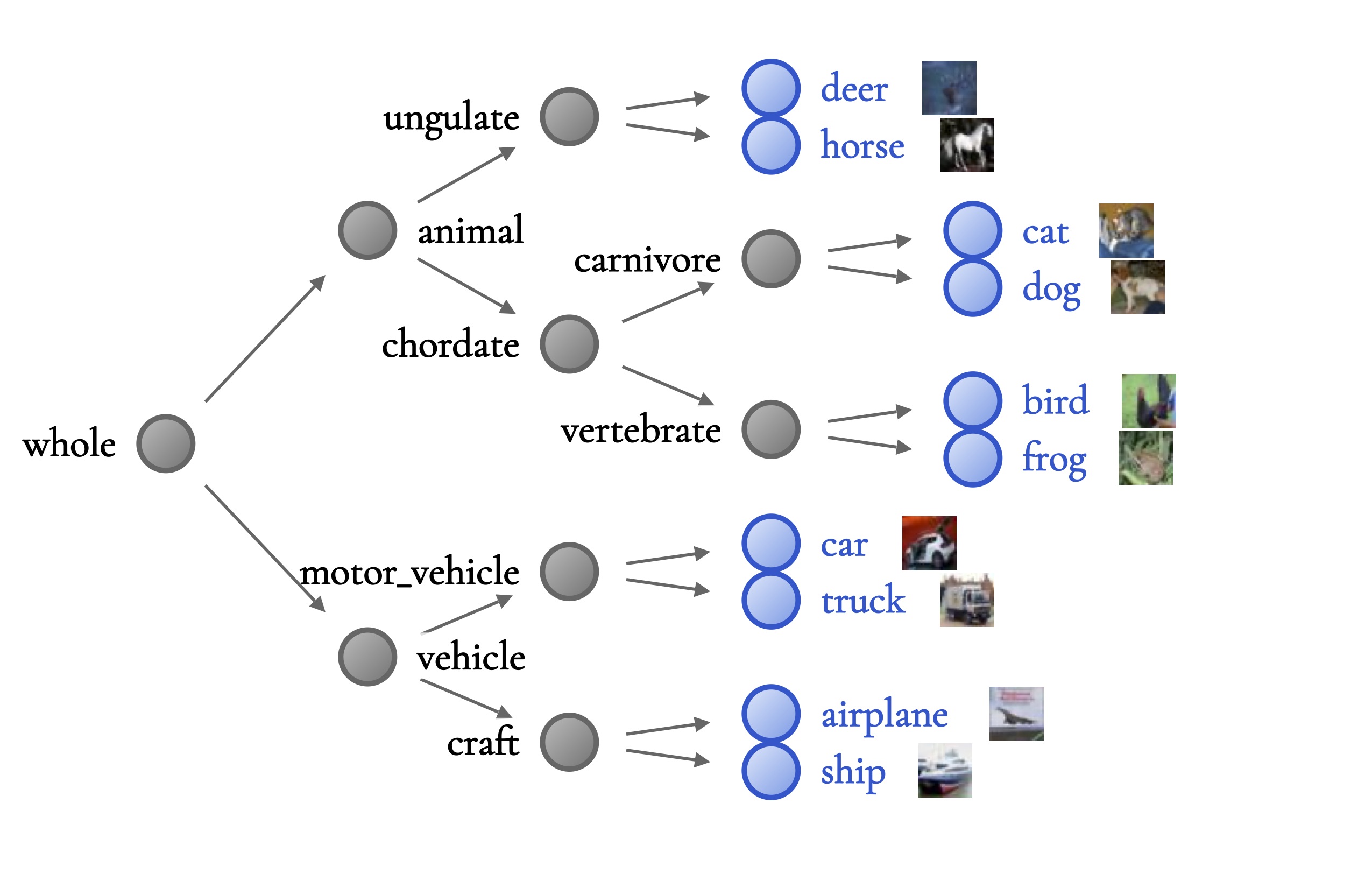}
\caption{A tree-shaped model, composed of several binary classifiers, designed for CIFAR-10 \cite{DecisionTrees}.}
\end{center}
\end{figure}

Other classifications, such as tertiary classifications, were tested, but following previous work, this classification schema performed the best and thus was utilized during testing.

\subsection{Input stability}
Although the noise necessary to guarantee differential privacy is often added during gradient descent, we hypothesize that, in some cases, adding noise directly to training data can be more effective. Different training data points vary in many different ways, but only some of this variation is useful for machine learning models. Therefore, if we can determine preprocessing methods that remove some of this unnecessary variation without significantly degrading model performance, this strategy may be more effective by decreasing the magnitude and/or dimensionality of the noise that must be added to the samples. If some measure improves the stability of the preprocessed data without substantially harming the model's performance, then that measure will make applying differential privacy carry less of a performance cost.

\subsection{Linear regression}
Recent testing has focused on improving the stability of the linear regression. The underlying technique that enables this is the use of pretrained models to transform the images in CIFAR-10. To move testing to the linear regression, CIFAR-10 images were preprocessed using large pretrained models developed on larger datasets. By training large models on image datasets like CIFAR-100 and imagenet, their convolutional layers can be extracted and applied as transforms to the CIFAR-10 dataset, resulting in a final product that transforms the image into a 1d vector with features highlighted by the use of these convolutional layers.  

This transform was tested through two major pre-trained models, trained on either CIFAR-100 or Imagenet. Before testing, hyperparameter optimization was conducted through single point removal, fixing the dataset, and removing one random point for varying dataset sizes ranging from 1,000 to 20,000 samples.  Full gradient clipping was also used, similarly to the Resnet20, and several thresholds were also tested. The values used for full gradient clipping were 0.25 and 0.05 to test the more drastic values of gradient clipping as we already had results from higher values.

For deterministic testing, both transforms were tested on randomly chosen subsets of 1,000, 5,000, 10,000, and 20,000 samples using a full batch gradient with Nesterov momentum for optimization. Percent divergence l2 norm along with average accuracy were tracked across 75 epochs. 8-10 different models, depending on the experiment, each with its own randomly-chosen subset of CIFAR-10, were used simultaneously.

Regularization was also used as another option for increasing stability. Using PyTorch's built-in weight decay, regularization was tested for all the training values above with both the Imagenet and CIFAR-100 transforms.

\subsubsection{Group-Sample Gradient Clipping}
\textbf{Motivation: Per-Sample Gradient Clipping}

A common method of improving model stability and sensitivity is to clip the gradient of each individual sample during training, ensuring it does not exceed a specific threshold. This ensures that each sample can only have a limited, well-defined effect on the model, giving a strong guarantee of high stability and low sensitivity. This clipping method is a vital part of state-of-the-art privacy frameworks for machine learning, such as the Python library Opacus. However, since we are using PAC privacy while Opacus uses differential privacy, we only need to improve stability, and are not bound to minimizing sensitivity.
\newline
\textbf{Implementation}

Since sensitivity is not an issue, our goal is to preserve the stability-improving properties of per-sample gradient clipping while simultaneously reducing its negative impacts on accuracy. To achieve this goal, we replace the practice of clipping the gradients of individual samples in a batch with clipping the gradients of small groups in a process similar to microbatching. If a hypothetical model uses a batch size of 1,000 samples, we divide that batch into smaller groups of samples (the size of which varies dramatically) and then clip the gradient of each group before averaging the gradients together and updating the model based on the output. We theorize that this will provide much of the stability improvement we would see from clipping individual samples (although its sensitivity will necessarily be worse by a factor equal to the group size). At the same time, group-sample clipping degrades the gradients from individual samples less, providing an end result closer to the non-private state of the art (or, equivalently, allowing clipping to be more aggressive while providing comparable results). In other words, we hypothesize that, as we grow the size of the group being clipped (which is, by definition, equal to 1 in per-sample clipping), the stability decrease (and a corresponding increase in noise added) will be more than counterbalanced by the accuracy increase.
\subsubsection{Dynamic Baseline Testing}

Instead of adding noise at the end of training, another option we have explored is adding noise periodically during the training process. To do this, we calculate the model's divergence after a set number of epochs, privatize the partially-trained model, and then continue training. Since we calculate divergence by comparing several separate models trained on different data, this means that we periodically reset all models to equal an arbitrary model (note that we do not choose the best-performing model, as this would introduce a positive bias to our results that would not occur during actual use). This process can be beneficial because divergence tends to grow super-linearly over time, meaning that total divergence can be minimized by "restarting" the training process repeatedly, allowing us to add small amounts of noise to the model several times during training, instead of adding a large amount of noise only at the very end.

\subsection{Combinations}
To test the effectiveness of all discussed stability methods, we combine all compatible stability models and train a Resnet 20 on the dataset, measuring stability and then adding noise through the eigenvalue method to create optimal anisotropic noise. Similarly, the Resnet 20 is trained for 100 epochs, and pruning, clipping, and regularization are applied. Additionally, linear regression was implemented with pruning, clipping, and regularization following a data transformation and 50 epochs of training. We use a regularization value of $10^{-2}$, due to its lower deviation and minimal accuracy loss. We pruned 30\% of the model before training and clipped the entire batch to 0.5. These values are consistent across both the resnet 20 and linear regression. 
\pagebreak
\section{Results}

\subsection{Baselines}
\subsubsection{Full-Batch Gradient}

When testing for stability, we first determined the inherent divergence of several models. Starting with a one-layer linear model of only 10,000 parameters and using full batch gradients, a 0.07 deviation l2 norm and 0.48 square root sum of eigenvalues were observed, along with a 3-4 percent increase in accuracy from an average of 41 to 44 percent. Having established a baseline when starting with a pretrained base, this was further extended to the Resnet20 model. When training several Resnet20 models with 25,000 samples from a pre-trained starting point with 5,000 samples, a roughly linear trend was observed in the first 50 training epochs when aiming for stability. This yielded an increase of around 10 percent in accuracy from 60 to 70 percent. 
\begin{wrapfigure}[28]{l}{7.1cm}
\begin{center}
\includegraphics[width=7cm]{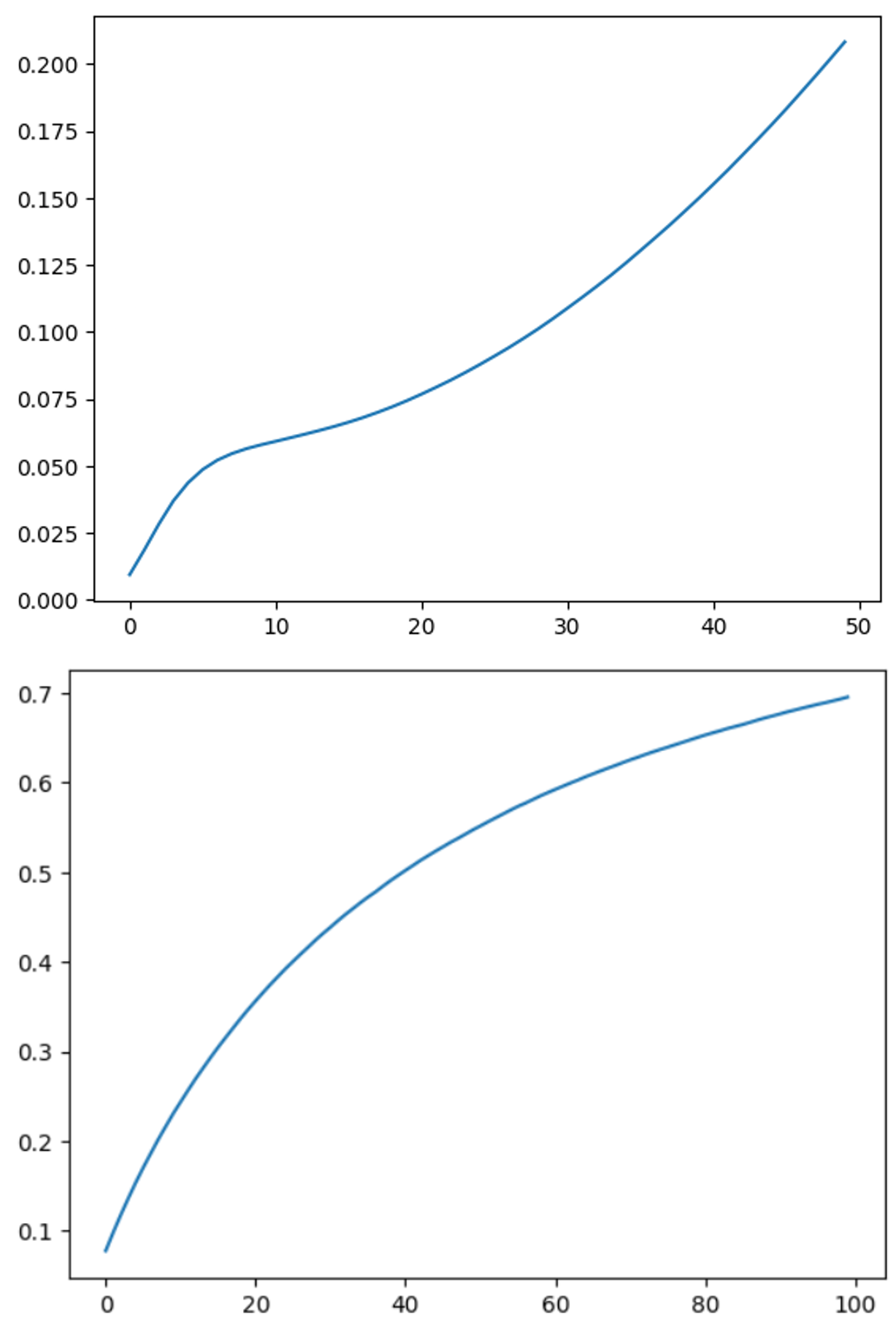}
\caption{Percentage l2 norms over training time (in epochs), with full batch pictured above and small batch picture below}
\end{center}
\end{wrapfigure}
When optimizing more for accuracy with smaller batch sizes, a much higher l2 norm was reached, with a curve that flattened off around a l2 norm 70 percent of the original model. However, this produced a 16 percent increase in accuracy, with the l2 norm of this approach being over double that of the high batch size at 50 epochs and yielding around twice the accuracy increase. While a full batch gradient yields far better stability, the longer training times and potential continuous linear trend mean that the divergence may overtake that of a smaller batch size.

\subsection{Experimentation}
\subsubsection{Layer Freezing}
After conducting layer freezing, the overall result was a low eigenvalue square root sum and low deviation l2 norm, being around 0.02 when calculated on the unfrozen linear layers used. This meant that the noise created as a result of the eigenvalues used had a negligible impact of around 0.3 percent on the overall accuracy of the model, but also meant that the accuracy increases as a result of training on a larger data subset was largely inconsequential, with an accuracy increases of only around 1-2 percent. This makes the overall notion of unfreezing layers unlikely to be optimal due to a minimal accuracy increases that a subsequent decrease in stability can easily negate. For this approach to be worth considering, the pre-trained model used as the base must already have high accuracy, which can be preserved through layer freezing. However, this isn't likely or optimal in a real-life setting, given that as much value as possible should be extracted from any private data instead of relying solely on available public data. 

\subsubsection{Pruning and Gradient Clipping}
Pruning models after they completed training yielded worse results than pruning before. While pruning 20 percent of parameters both before and after training led to a deviation of 57 percent and a 14 percent increase in accuracy, pruning half of the parameters beforehand caused only a 55 percent deviation while simultaneously increasing accuracy by 15 percent, making it more viable than pruning after training.

While pruning led to a decrease in the l2 norm of the deviation over just using the pre-trained model, gradient clipping produced far more promising results. Clipping the l2 norm of each batch gradient to 0.5 led to an increase in accuracy from 67 percent to 80 percent, while the percent deviation l2 norm only reached 20 percent, improving upon the 10 percent increase from the base model while maintaining the lower deviation l2 norm. With gradient clipping, the trend appeared linear instead of quadratic, likely due to the minimized impact of outliers. 

\subsection{Linear Regression}
After conducting data transformation, the resulting accuracy of the linear regression was increased to around 70 percent with the CIFAR-100, and around 90 percent with the Imagenet transform on average. These accuracies can be used as a baseline to compare other results to, since they are the theoretical best accuracy that could be attained with a linear model (due to the fact that linear models are inherently convex). 
\subsubsection{Point Removals}

During initial testing, a baseline of 1 point removal for the 1,000, 10,000, and 20,000 was established and measured in absolute deviation l2 norm. This and additional hyperparameter testing are compiled in the table below. Overall, training for longer has diminishing returns throughout all dataset sizes, and the 10 times increase in learning rate affects both the 1,000 and 10,000 deviation norms by roughly 3 times, while affecting the 20k by over 10 times. The increased divergence from utilizing the higher rate was mostly outweighed by the accuracy increase in 10k, but the especially large increase of over 10 times for 20k makes it less feasible. Due to the high accuracy of 10k approaching 20k and its relatively excellent l2 norm, it was used for further testing with single-point removal. Due to the rapid training and high accuracy of the 0.1 learning rate for the linear regression, it was utilized for future testing. 
\begin{wraptable}{l}{12.4cm}
\begin{tabular}{ |p{2cm}|p{1.25cm}|p{1.75cm}||p{3.5cm}|p{1.5cm}|  }
 \hline
 \multicolumn{5}{|c|}{Accuracy and Deviation for Various Hyperparameter Values} \\
 \hline
Number of Samples&Epochs&Learning rate& l2 Norm Deviation &Accuracy\\
 \hline
1,000& 100& 0.01& 0.025    &71.1\\
1,000& 50& 0.1& 0.067   &73.4\\
1,000& 100& 0.1& 0.096   &73.7\\
 \hline
10,000& 100& 0.01& 0.002    &71.9\\
10,000& 50& 0.1& 0.007   &75.7\\
10,000& 100& 0.1& 0.011  &77.0\\
 \hline
20,000& 100& 0.01& 0.0004   &75.9\\
20,000& 100& 0.1& 0.005   &77.4\\
 \hline
\end{tabular}
\end{wraptable}
For the 10,000-point dataset, removals of 10, 100, 1,000, 2,500, and 5,000 points were tested on the model to measure its resistance to more drastic changes and compare it to the expected increases in divergence. Naturally, you would expect the variance to be multiplied by 10 times when removing 10 points instead of 1, and so on. The absolute deviation l2 norms produced from these removals are as follows: 0.04 for 10 points, 0.14 for 100 points, 0.46 for 1000 points, 0.80 for 2500 points, and 1.37 for 5000 points.
\pagebreak
\begin{wrapfigure}{l}{9.1cm}
\begin{center}
\includegraphics[width=9cm]{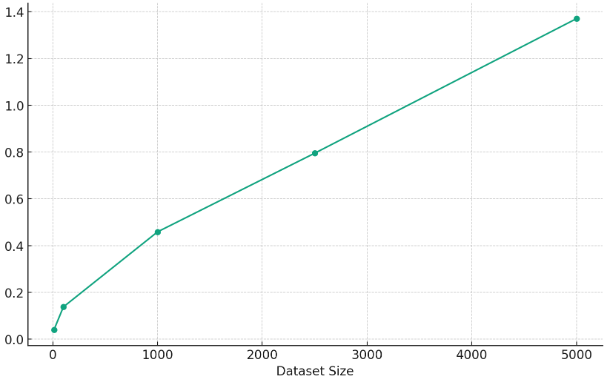} 
\caption{Absolute deviation for points removed from a 10,000-point dataset. The number of points removed is on the X axis, and deviation is on the Y axis}
\end{center}
\end{wrapfigure}

As expected, we can see in Figure 5 that a linear trend emerges between the chosen points, plateauing from the steeper trend of the first few data removals. While the 0.04 deviation norm exceeds the expected 0.02 (10 times our 1-point baseline of 0.002), the other data points are lower than expected despite having a linear trend. It can also be noted that the deviation from 5,000 random points removed from 10,000 surpasses that from 10,000 points sampled completely at random, likely due to the smaller, albeit shared sampling size, which could lead to more overfitting on specific traits. As for accuracy, the decreases in accuracy also roughly follow a linear trend, going from 77.0 to 76.6 with the removal of half the dataset, matching the similarly linear increase of the deviation l2 norm.

\subsubsection{Whole Batch Gradient Clipping}
For the results of whole batch gradient clipping with thresholds of 0.25 and 0.05 using 1,000, 10,000, and 20,000 sample datasets with 1 point removed, the resulting deviation l2 norms using a learning rate of 0.1 and training for 100 epochs are plotted in the table below.

\begin{wraptable}{l}{10.9cm}
\begin{tabular}{ |p{4.5cm}||p{2.5cm}|p{2.5cm}|  }
 \hline
 \multicolumn{3}{|c|}{Deviation l2 Norm for Whole Batch Gradient Clipping} \\
 \hline
Dataset size& 0.25 clipping & 0.05 clipping\\
 \hline
1,000 & 0.01    &0.008\\
 \hline
10,000& 0.009  &0.004\\
 \hline
20,000& 0.005   &0.005\\

 \hline
\end{tabular}
\end{wraptable}
It can be noted that the use of clipping heavily affected the sample size of 1,000 points, cutting down on the deviation by almost 100 times compared to the same hyperparameters without clipping. However, the 10,000 and 20,000 sample sizes were much less affected, producing little to no change in deviation. In contrast, the accuracy of these results had little to no variance, with accuracies dropping by 2 percent for all data point values when compared to training without clipping. Overall, this form of whole batch gradient clipping seems to disproportionately benefit smaller sample sizes, likely because the use of whole batch gradient for larger sample sizes allows the regression not to have any outliers that need clipping, which makes their use redundant.

\subsubsection{Transform Efficiency}
After the baselines of point removal, testing was conducted on data transformed by pretrained models across sets of 1,000, 5,000, 10,000, and 20,000 samples. The results for both percent deviation l2 norm and accuracy across 75 epochs can be seen below in Figures 6 and 7. The four graphs in Figure 6 each correspond to a different dataset size, and the blue lines are training results with the CIFAR-100 transform while the orange lines are results with the Imagenet transform.

\pagebreak

\begin{wrapfigure}{l}{11.1cm}
\begin{center}
\includegraphics[width=11cm]{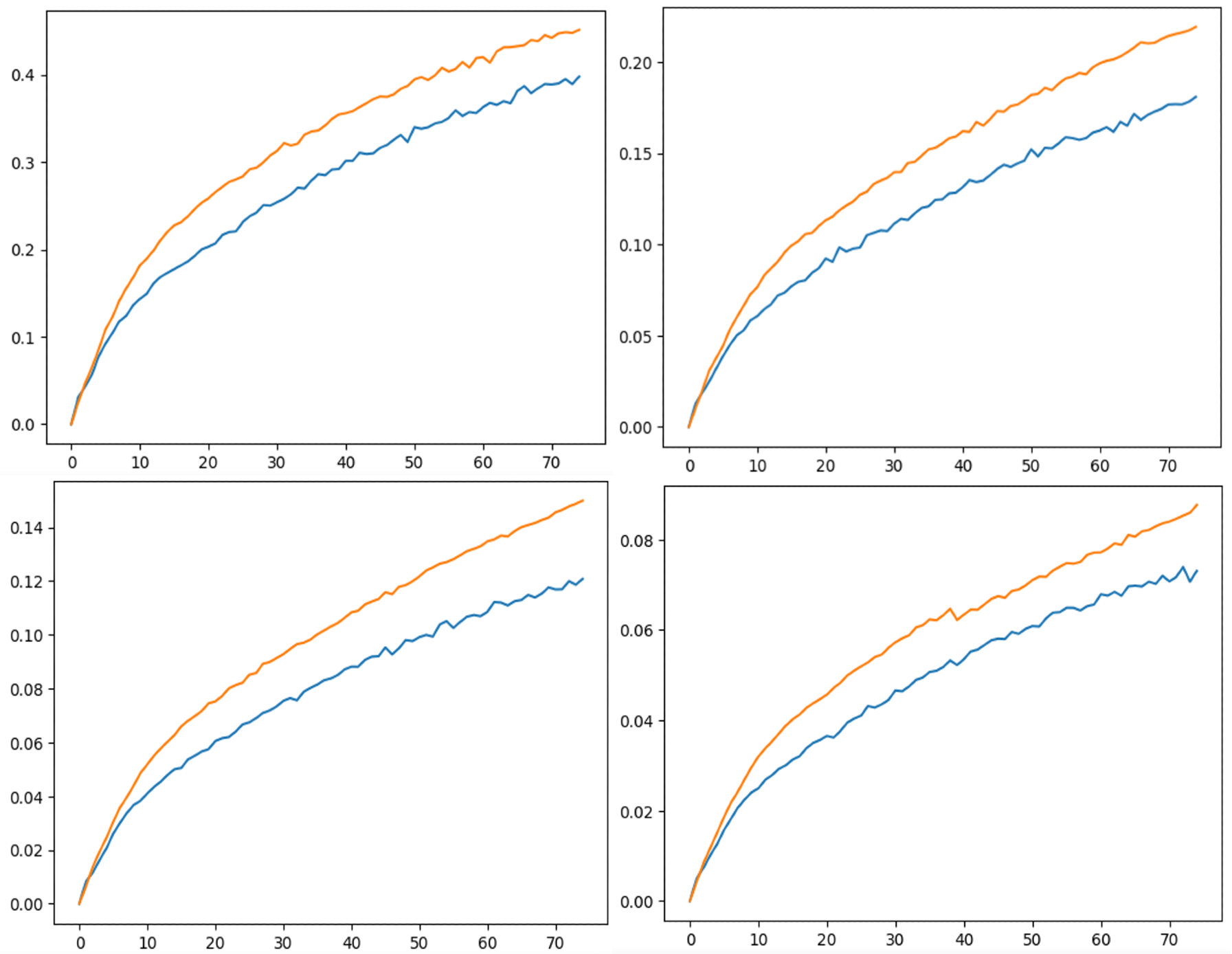} 

\caption{Deviation l2 norm of the CIFAR-100 and Imagenet transforms tracked across 75 epochs for 1,000, 5,000, 10,000, and 20,000 randomly subsampled datapoints. 1,000 points in the top left, 5,000 in the top right, 10,000 in the bottom left, and 20,000 in the bottom right}
\end{center}
\end{wrapfigure}
 
The same pattern emerges for all data values, with the l2 norms from data preprocessed with Imagenet models consistently trending above the CIFAR-100 norms. Both lines start to converge and increase more slowly as time passes, but the number of epochs necessary to converge completely is likely too large to be feasible, especially when l2 norms continue climbing. This result is somewhat expected, as the accuracies produced by the Imagenet preprocessing all surpass those produced by CIFAR-100 by an average of 16.7 percent across all data values. However, the accuracy increases between sample sizes while using the same transform are far less pronounced, with the accuracy for 1,000 data points on CIFAR-100 being only 3.5 percent lower than 20,000 data points on CIFAR-100 and 1.9 percent lower on Imagenet for the corresponding values. As an example, the accuracy graph for 10,000 in Figure 7 below shows that accuracy jumps and approaches the local minimum within the first 10 epochs, then converges. An optimal training value may be 50 epochs instead of 75, which could reduce the divergence l2 norm while minimizing the accuracy loss since the last 25 epochs are somewhat redundant.  

\begin{figure*}[b]
\begin{center}
\includegraphics[width=7cm]{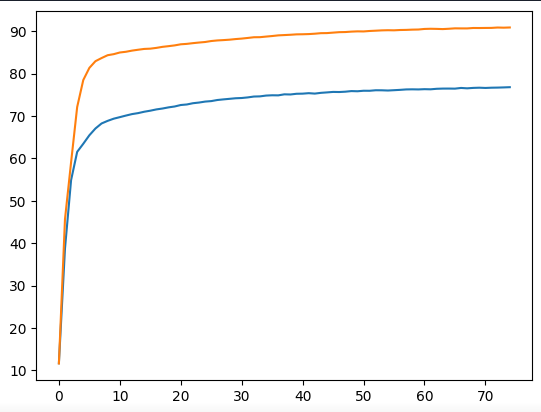}
\caption{Average accuracy of 10,000 random data points using both CIFAR-100 and Imagenet transforms, with Imagenet transforms being the orange line and CIFAR-100 being the blue line}
\end{center}
\end{figure*}

\pagebreak
\subsubsection{Regularization}
The regularization results are displayed below in a similar fashion to the random subsets across 4 graphs, each for a separate data point value. We have tested with both the Imagenet and CIFAR-100 preprocessing and they behaved similarly; only the CIFAR-100 results are discussed here. The model without any regularization is tracked with the blue line, the model with a regularization term of $10^{-3}$ is red, a term of $10^{-2}$ is green, and a term of $10^{-1}$ is orange. 

\begin{wrapfigure}{l}{11.1cm}
\begin{center}
\includegraphics[width=11cm]{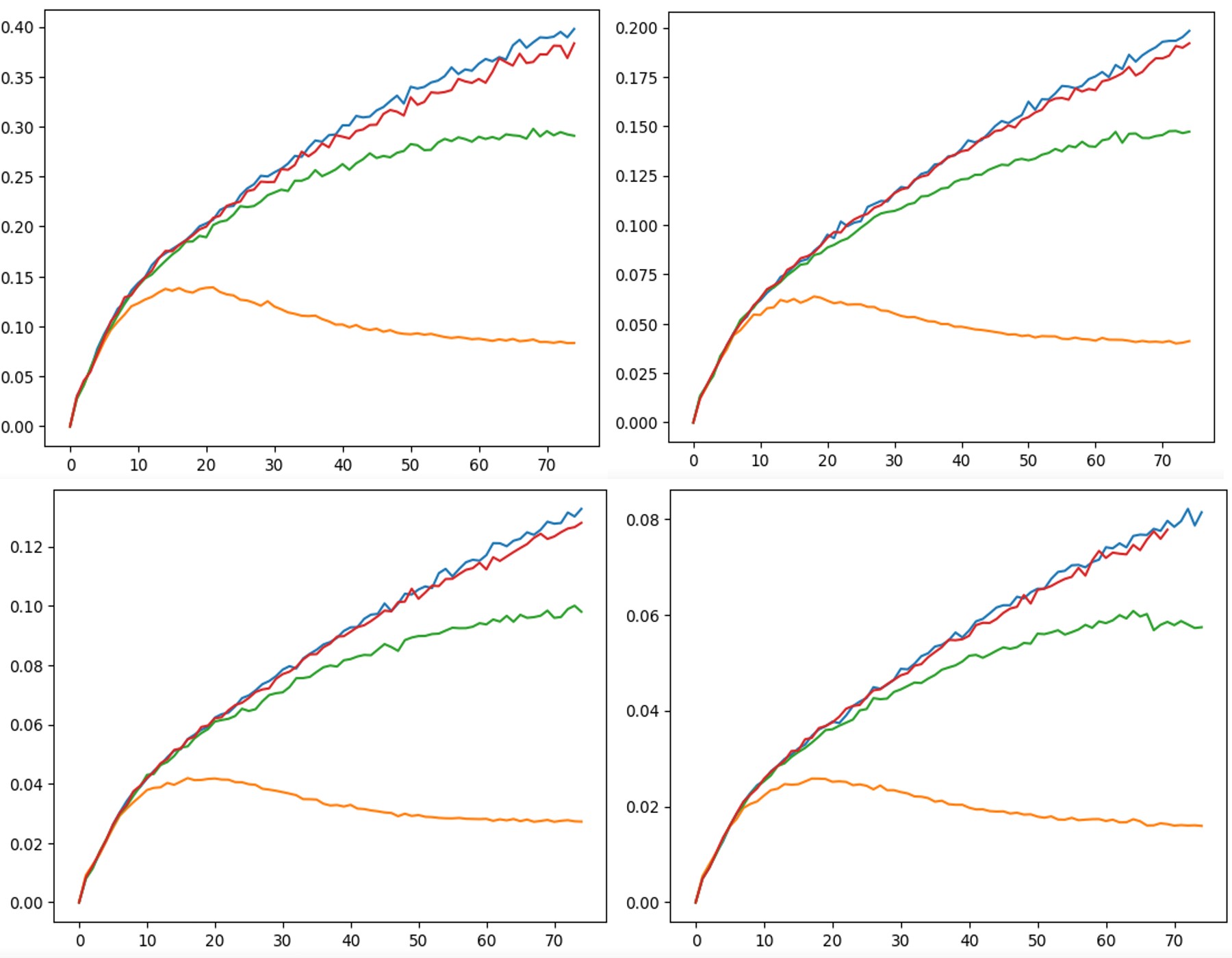}
\caption{Deviation l2 norm of various regularization values tracked across 75 epochs for 1,000, 5,000, 10,000, and 20,000 randomly subsampled datapoints. 1,000 points is in the top left, 5,000 in the top right, 10,000 in the bottom left, and 20,000 in the bottom right}
\end{center}
\end{wrapfigure}

Across all data points, the regularization value of $10^{-3}$ mostly mirrors the original without any regularization, the regularization value of $10^{-2}$ performs significantly better, and the regularization value of $10^{-1}$ allows the linear regression to converge completely. However, while the regularization values of $10^{-3}$ and $10^{-2}$ attain similar accuracies to the original, the regularization value of $10^{-1}$ causes a decrease in accuracy. Figure 9 shows the accuracies for various regularization strengths.

\begin{figure*}[b]
\begin{center}
\includegraphics[width=6cm]{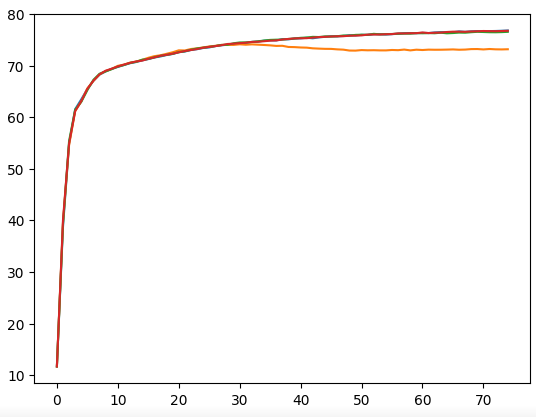}
\caption{Average accuracy of 10,000 random data points with various regularization values. The value of $10^{-1}$ for regularization is the orange line that diverges}
\end{center}
\end{figure*}

It can be noted that while lower regularization values achieve extremely similar accuracies to the base model, only the value of $10^{-1}$ deviates from the rest, converging at a lower accuracy. We can also note that the peak of its accuracy occurs when the deviation l2 norm of the regularization value peaks, exemplifying the accuracy and stability tradeoff seen before. While final accuracy falls by about 5 percent, the extreme benefits in stability may in some cases outweigh the minor loss in accuracy if strong privatization is necessary. 
\newline
\newline
Overall, regularization provides another effective approach to stabilization, acting in some ways similarly to gradient clipping by normalizing all gradients to a certain extent. 

\subsubsection{Group-Sample Gradient Clipping}
We tested group-sample gradient clipping with several different group sizes. As in previous experiments, we trained a simple one-layer linear model. As before, our dataset was created by taking CIFAR-10 and preprocessing it with the convolutional layers of a separate model trained on CIFAR-100. Each model was trained on 20,000 randomly sampled data points from the train set and tested on the entire test set. We carried out trials at 4 different group sizes. Each trial was carried out with 8 independently trained models to get a better measurement of stability. In all trials, we used simple, full-batch gradient descent with a learning rate of 0.1, and group norms were clipped to a maximum magnitude of 1.0. The largest group size had a minimal effect, with models attaining similar accuracy and stability to models trained without any clipping (less than a 1\% change in accuracy and negligible change in divergence).

\begin{wrapfigure}[32]{l}{8.5cm}
\begin{center}
\includegraphics[width=8.0cm]{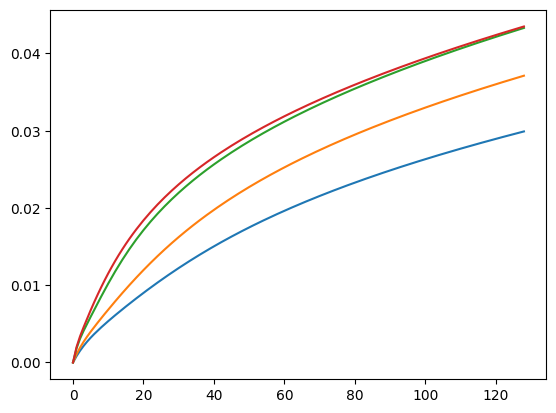}
\includegraphics[width=8.0cm]{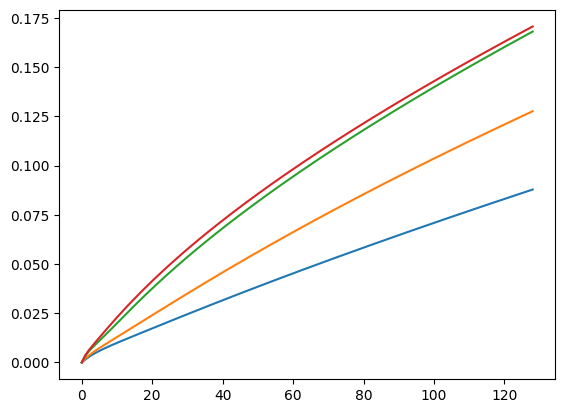}
\caption{Percentage (above) and absolute (below) divergence of models trained with group-sample gradient clipping. Blue, orange, green, and red lines correspond to group sizes of 10, 25, 100, and 400, respectively.}
\end{center}
\end{wrapfigure}

As we expect, models trained with smaller groups diverge significantly slower (i.e. have significantly better stability) than their counterparts with large groups. Stability is improved by a factor of nearly two when the group size is lowered from 400 samples to 10. At the same time, accuracy only drops by 3\%, from 73.5\% to 70.5\%. The generally low accuracy in both cases is due in part to the models' small size and in part to other simplifications such as the unsophisticated gradient descent algorithm (our limited testing suggests that using momentum can, in this specific case, improve results across the board by approximately 2 percent). The trends in divergence shown in Figure 10 continue for several hundred epochs beyond the end of the graph, with divergence continuing to increase while accuracy stays almost the same. If group-clipping strength increases further, divergence continues to decline, but accuracy also begins falling faster, making the overall tradeoff no longer beneficial.

\subsubsection{Dynamic Baseline Testing}

To test the viability of potentially adding noise to the model periodically during the training process, we train 16 regression models on 10,000 random samples from CIFAR-10 (preprocessed in the same way as previous experiments) for 16 epochs. After those 16 epochs, we re-initialize all 16 models to be equal to one of them (chosen arbitrarily), re-select our sets of 1,000 random datapoints, and train for 16 more epochs. This is repeated 8 times, for a total of 128 epochs with 10,000 datapoints each. Our results are shown in the figure below.

\begin{figure}
\begin{center}
\includegraphics[width=11.0cm]{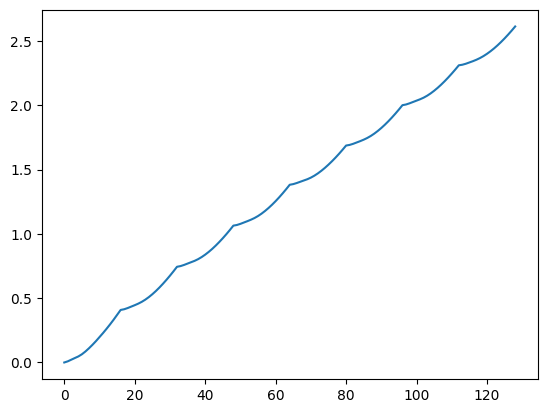}
\caption{Absolute divergence of models trained with regular re-initialization. Note the 8 distinct segments, corresponding to the 8 separate sets of 16 epochs during training.}
\end{center}
\end{figure}

Overall, this method leads to a significantly higher total divergence than alternative methods, likely due to the fact that the subset of CIFAR-10 we train on changes regularly, causing the model to vary more than it has to in order to achieve its goal. Despite, or perhaps because of this very high divergence, models trained this way perform fairly well, attaining a test accuracy of 77\%.

\subsection{Combinations}
Overall, when combining the several methods used, the accuracy of the trained model somewhat suffers, but the divergence also decreases consistently, leading to minimal accuracy loss from the anisotropic noise added. The Resnet 20, after 100 epochs, was able to achieve an accuracy of 76.1\%, with a deviation l2 norm of 12\%, much lower than any individual method. Adding noise through eigenvalues then results in an accuracy decrease of 3.6\% to drop accuracy to 72.5\%. For the linear regression, with the CIFAR-100 transform, the regression achieves a similar accuracy of 75.2\%, but a much lower deviation l2 norm of 1.6\%, leading to the lower accuracy decrease of \%1.7 to achieve 73.5\%. With the higher ending average, the linear regression with a data transform seems to be the better option, with its drawback being the requirement of abundant similar data to create such a transform. Compared to the Resnet20 trained with batches and no augments for 100 epochs, these models perform much better, with the basic Resnet suffering a large loss of 31.9\% due to its extremely high deviation, thus resulting in a far inferior performance. 

\section{Conclusion}
In this work, we studied various techniques to optimize the stability of both convex and non-convex machine learning models. Of the techniques we tested, the most effective included pruning, pretrained data transforms, regularization, and group-sample gradient clipping. These methods significantly reduced model divergence over time, thereby improving model stability with minimal harm to test accuracy and allowing for more effective model privatization with a smaller overall decrease in accuracy. Overall, given the tradeoff between accuracy and stability, the pretrained data transform acts as the most efficient intervention of the methods we study in this paper, but it is not always possible to implement, due to the need for a public dataset similar to the private data being used. When a pretrained transform is not possible (due to lack of public data), gradient clipping was shown to be the next best choice. 

\section{Future work} 
For future work, continued work in the area of dynamic bases and data augmentation can be pursued. While the idea for data augmentation, augmenting data by 10 times for each model trained to increase stability and result in less noise added, has been outlined in theory, it has not been executed with concrete results. Due to this, any future work pursued would like to focus on this augmentation to reduce data needed to train models and increase stability during the training process, potentially combining this with various augmentation and privacy methods previously discussed. 

\section{Acknowledgements}

This work would not have been possible without the MIT PRIMES program and the assistance of our great mentor, Hanshen Xiao, who directed us to the resources we needed and helped us figure out the state-of-the-art in the field and which ideas to pursue.

\pagebreak

\printbibliography

\end{document}